\documentclass[sigconf,nonacm]{acmart}

\usepackage[T1]{fontenc}
\usepackage{soul}
\usepackage{comment, kotex}
\usepackage{graphicx}
\usepackage{bbm}
\usepackage{enumitem}
\usepackage{amsmath, color}
\usepackage{algorithm}
\usepackage{algpseudocode}
\usepackage{float}
\usepackage{xcolor}
\usepackage{amsfonts}
\usepackage{graphicx}
\usepackage{multirow}
\usepackage{comment}
\usepackage{caption}
\usepackage{hyperref}
\usepackage{orcidlink}

\AtBeginDocument{%
  }

\setcopyright{acmlicensed}
\copyrightyear{2018}
\acmYear{2018}
\acmDOI{XXXXXXX.XXXXXXX}

\acmConference[Conference acronym 'XX]{Make sure to enter the correct
  conference title from your rights confirmation email}{June 03--05,
  2018}{Woodstock, NY}
\acmISBN{978-1-4503-XXXX-X/18/06}

\newcommand{\fraName}{GDFlow}

\begin{document}

\title{GDFlow: Anomaly Detection with NCDE-based Normalizing Flow for Advanced Driver Assistance System}

\author{Kangjun Lee}
\orcid{0000-0002-0788-721X}
\email{gkdl677@g.skku.edu}
\affiliation{
  \institution{Department of Computer Science and Engineering\\ Sungkyunkwan University}
  \city{Suwon}
  \country{South Korea}}

\author{Minha Kim}
\orcid{0000-0002-3224-0610}
\email{sunshine01@g.skku.edu}
\affiliation{
  \institution{Department of Artificial Intelligence\\Sungkyunkwan University}
  \city{Suwon}
  \country{South Korea}}

\author{Youngho Jun}
\email{YoungHo.Jun@hyundai-kefico.com}
\affiliation{
  \institution{Hyundai-KEFICO}
  \city{Gunpo}
  \country{South Korea}}

\author{Simon S.Woo}
\authornote{Corresponding author.}
\orcid{0000-0002-8983-1542}
\email{swoo@g.skku.edu}
\affiliation{
  \institution{Department of Applied Data Science\\Sungkyunkwan University}
  \city{Suwon}
  \country{South Korea}}

\renewcommand{\shortauthors}{Kangjun et al.}

\begin{abstract}
For electric vehicles, the Adaptive Cruise Control (ACC) in Advanced Driver Assistance Systems (ADAS) is designed to assist braking based on driving conditions, road inclines, predefined deceleration strengths, and {user} braking patterns. However, the driving data collected during the development of ADAS are generally {limited and lack diversity. This deficiency leads to} late or aggressive braking for different users.
{Crucially, it is necessary to effectively identify anomalies, such as unexpected or inconsistent braking patterns in ADAS, especially given the challenge of working with unlabelled, limited, and noisy datasets from real-world electric vehicles.}
In order to tackle the aforementioned challenges in ADAS, we propose \underline{\textbf{G}}raph Neural Controlled \underline{\textbf{D}}ifferential Equation Normalizing {\underline{\textbf{Flow}}} (\textbf{\textit{\fraName}}), a model that leverages Normalizing Flow (NF) with Neural Controlled Differential Equations (NCDE) to learn the distribution of normal driving patterns continuously. Compared to the traditional clustering or anomaly detection algorithms, our approach effectively captures the spatio-temporal information from different sensor data and more accurately models continuous changes in driving patterns. Additionally, we introduce a quantile-based maximum likelihood objective {to improve the likelihood estimate of the normal data near the boundary of the distribution, enhancing the model’s ability to distinguish between normal and anomalous patterns.}
We validate \fraName~using real-world electric vehicle driving data that we collected from Hyundai IONIQ5 and GV80EV, achieving state-of-the-art (SOTA) performance compared to six baselines across four dataset configurations of different vehicle types and drivers. Furthermore, our model outperforms the latest anomaly detection methods across four time series benchmark datasets. Our approach demonstrates superior efficiency in inference time compared to existing methods.
\end{abstract}

\ccsdesc[500]{Mathematics of computing~Time series analysis}
\ccsdesc[500]{Computing methodologies~Anomaly detection}

\keywords{Advanced Driver Assistance Systems, Multivariate Time Series Anomaly Detection, Normalizing Flow, Neural Controlled Differential Equations}

\maketitle

\section{Introduction}
The Adaptive Cruise Control (ACC) in the Advanced Driver Assistance System (ADAS) for electric vehicles assists in {automatic} braking based on driving conditions, road inclines, and the driver's deceleration patterns without having them to manually press the {pedal}~\cite{segata2013automatic, chengqun2023novel, bengtsson2001adaptive}. 
However, the driving data collected during the vehicle development process is quite limited and lacks the diversity of various driving patterns. Due to this limitation, the ACC often relies on predefined deceleration strengths, {which are generally set} to three levels: strong, normal, and smooth deceleration. However, such a high level discrete categorization can lead to potential discomfort for drivers, as it may not accurately reflect individual driving preference and behavior~\cite{adornato2009characterizing, li2022unsupervised}. As the intensity and sensitivity of automatic braking in ADAS is determined based on the {limited} data collected during development, the system may operate too late or too aggressively for different users~\cite{lv2018hybrid},  which can be problematic. 

In the existing {ACC}~\cite{jun2022machine}, DBSCAN was used {to} detect anomalies in the collected driving data to enhance driving stability by identifying consistent deceleration patterns. However, classical clustering algorithms such as DBSCAN are highly influenced by {variations} in data density,
{making it subject to high sensitivity w.r.t different hyperparameters.}
Furthermore, DBSCAN struggles under different traffic conditions, being vulnerable to high sensitivity under hyperparameters~\cite{akccelik2001acceleration, borg2023ergo, li2022unsupervised}.
Moreover, the real-world ADAS in electric vehicles must handle {complex and noisy} multivariate time series (MTS) sensor data~\cite{hajek2013workload}. While several clustering methods were introduced to detect, they face challenges due to high sensitivity to hyperparameters. Also, deep learning-based methods often use supervised approaches~\cite{gornitz2013toward, jia2019anomaly, castellani2020real} {that require annotation}, which is labor-intensive~\cite{siegel2020industrial} and susceptible to label noise.

In order to tackle the aforementioned challenges, we adopt unsupervised one-class classification (OCC) approaches to effectively perform anomaly detection by learning only normal data patterns. In particular, we explore a {Normalizing Flow} (NF)-based model, assuming that anomalous data {is located in} the low-density regions of the data distribution. We find that {NF} can effectively model the complex underlying normal data distributions. NF-based models have been proposed for anomaly detection in real-world MTS datasets, using {Graph Neural Networks} (GNN) for multiple entity dependencies and RNN models for temporal information~\cite{zhou2023detecting, dai2022graph}. However, these approaches are limited by the RNN's capacity to model complex and continuous temporal dynamics, whereas {Neural Controlled Differential Equations (NCDE)} are better suited to model {temporal variations} and extrapolate {to unseen future data.} Thus, we propose \fraName, which leverages NCDE-based NF to learn the distribution of normal driving patterns and effectively capture both spatial and temporal dependencies. Additionally, we employ a graph structure learning module that incorporates relational information from sensors used in {the ACC} of Hyundai vehicles.

Another significant challenge in ADAS is the evolving nature of user patterns as driving data accumulates over time, which complicates the modeling of driving behaviors.
{Inspired by recent research on anomaly detection using adaptive graph generation~\cite{guo2019attention, choi2022graph}, we apply Adaptive Graph Generation (AGG) to capture spatio-temporal information from sensor data and integrate it with the NF model for anomaly detection.}
Furthermore, we utilize the quantile function to optimize the likelihood estimates of normal data near the {boundary of the distribution}, enhancing the separation between normal and anomalous data~\cite{taghikhah2024quantile}.
To validate our GDFlow, we collected the real-world user driving dataset from Hyundai IONIQ5 and GV80EV electric vehicles and evaluated our method on eight different configurations for cars and drivers. Our method outperforms nine baselines, achieving the SOTA performance. In addition, our model performs better than the latest anomaly detection methods across four popular time series anomaly detection benchmark datasets.
Our research contributions are summarized as follows:

\begin{itemize}
    \item In ADAS, we are the first to propose the anomaly detection method based on NF with NCDE to simultaneously capture temporal and spatial {information} and effectively model continuous user driving pattern changes.

    \item Additionally, we introduce a quantile-based maximum likelihood objective to better learn the normal driving patterns of users near the boundary of the distribution.
    
    \item We collected the real-world braking patterns and deceleration datasets from Hyundai IONIQ5 and GV80EV. {We also} evaluated our methods in addition to popular benchmark datasets, achieving the SOTA performance.  
\end{itemize}

We have integrated and tested GDFlow in real-world vehicle systems and plan to deploy GDFlow in Hyundai Genesis GV90 by March 2026.

\section{Related Works}
\subsection{Clustering Algorithm for Electric Vehicles}
In the domain of electric vehicles, clustering algorithms have been used across various applications. Xie et al.~\cite{xie2018distribution} discuss a methodology for clustering driving data using DBSCAN to separate {high density} regions from noise, which is useful for identifying meaningful patterns in braking data. This approach is advantageous compared to traditional methods such as {K-means} since DBSCAN does not require predefining the number of clusters and can handle arbitrary shapes and noise effectively. Moreover, Straka et al.~\cite{straka2019clustering} explore multiple clustering algorithms, including K-means, DBSCAN, and agglomerative hierarchical clustering, to analyze usage-related segments of electric vehicles. Their approach aims to identify patterns in charging behavior and other usage metrics of charging stations, providing insights into optimizing charging infrastructure and vehicle utilization. However, classical clustering algorithms such as DBSCAN are inherently sensitive to hyperparameters such as radius value and the minimal number of points~\cite{an2023strp}.
To address this issue, STRP-DBSCAN~\cite{an2023strp} incorporates temporal attributes and uses parallel processing and an autotuning mechanism for DBSCAN parameters using deep reinforcement learning, enhancing clustering accuracy. In our case, traditional DBSCAN-based clustering algorithms also face significant challenges due to hyperparameter sensitivity, making it difficult to be realizable in real-world vehicles. Therefore, we focus on tackling this problem as an unsupervised anomaly detection from clustering algorithms.

\subsection{Deep Clustering Network}
Many studies enhance clustering algorithms using deep learning models, specifically autoencoders, to generate better representations of clustering algorithms. Lu et al.~\cite{lu2022improved} {employ} an autoencoder to extract latent features, assigning pseudo-labels through initial clustering. Reliable samples are selected for further training in a CNN, filtering out unreliable samples and improving clustering accuracy. Similarly, Li et al.~\cite{li2021deep} {use} an autoencoder-enhanced clustering method for anomaly detection in image data, iterating between hypothesizing normal candidate subsets and representation learning. The reconstruction error from the autoencoder serves as a scoring function to assess normality. For time series data, DTCR~\cite{ma2019learning} integrates a Seq2Seq model with a K-means clustering objective and a temporal reconstruction loss to generate cluster-specific representations. Further, Aytekin et al.~\cite{aytekin2018clustering} enhance clustering by adding an $l_2$ normalization constraint on autoencoder representations, making feature vectors more separable and compact. Hence, deep clustering networks have been used {to cluster} data in ACC. However, we take a completely different and new approach using {an unsupervised} anomaly detection method to address the problem more effectively. We demonstrate the superiority of our model by comparing its performance with conventional deep clustering networks.

\subsection{Multivariate Time Series Anomaly Detection}
Generally, anomaly detection algorithms can be categorized into one of the following categories: reconstruction, association, and forecasting-based approaches. The reconstruction-based approach, THOC~\cite{shen2020timeseries}, uses a dilated RNN with skip connections to capture temporal dynamics across multiple scales. {Moreover}, M2N2~\cite{kim2024model} incorporates a test-time adaptation strategy to adjust model parameters during inference, handling distribution shifts between training and test data. The association-based approach, established by the Anomaly Transformer (AT)~\cite{xu2021anomaly}, uses an Anomaly-Attention mechanism to capture both prior-association and series-association discrepancies, allowing the model to distinguish normal and anomalous points.
Recently, {NF}-based models have been used {for anomaly} detection, assuming anomalies are in low-density regions. In particular, the GANF model~\cite{dai2022graph} integrates a Bayesian network to model causal relationships and a graph-based dependency encoder to capture inter-feature correlations. MTGFlow~\cite{zhou2023detecting} builds on this framework, introducing a dynamic graph structure learning module using self-attention and an entity-aware NF for entity-specific density estimation. However, user patterns can exhibit distribution shifts over time in real-world {datasets}, such as driving data.
{NCDE has been utilized to model continuous changes in the data distribution more naturally and better extrapolate unseen data.}
In particular, the STG-NCDE model~\cite{choi2022graph} integrates GNN with NCDE to capture inter-feature correlations and temporal dynamics, outperforming existing models in accuracy and robustness to irregular time series data and missing observations. Based on this, we develop the first model to apply NCDE and NF within a single framework to model the spatio-temporal distribution of real-world deceleration data continuously in the ADAS system.

\section{Background on ADAS and Dataset}
\subsection{Advanced Driver Assistance System (ADAS)}
The ADAS is a driver convenience feature that enables vehicle operation without the need to press the accelerator or brake pedals. It adjusts to driving conditions, road gradients, and the driver's driving style. The fundamental control of ADAS involves calculating the target acceleration based on the distance and relative speed of the vehicle ahead. This is achieved by combining the Constant Time Gap Policy {(CTG)} with the uniform acceleration motion equation. The equations for the {CTG} and the uniform acceleration motion are presented as follows:
\begin{equation}
   a_{CTG} = -\frac{1}{h}(\dot{\epsilon} + \lambda \delta),
\end{equation}
\begin{equation}
   a_{UAM} = \frac{{v_{front}}^2 - {v_{ego}}^2}{2 \epsilon},
\end{equation}
where $\epsilon$ represents the distance to the vehicle ahead, $\lambda$ is the control parameter, $h$ is the target time gap, and $\delta$ is the distance error. Moreover, $v_{front}$ and $v_{ego}$ refer to the speed of the vehicle ahead and the ego vehicle, respectively. The {CTG} offers the advantage of preventing traffic congestion for following vehicles. Meanwhile, the uniform acceleration motion equation adjusts the distance between vehicles to calculate the target acceleration.

However, if a vehicle equipped with ADAS executes automatic control with delayed braking or excessive deceleration, the driver may experience discomfort. To address these issues, various studies are being conducted to learn from the driver's deceleration data. The goal is to control the vehicle in a manner that closely resembles the driver's usual deceleration habits. Since driver braking data encompasses various deceleration scenarios, an algorithm for extracting stable deceleration data is required. This study proposes a method to extract data that represents the driver's decelerating habits through deep learning methods.

\subsection{Deceleration Dataset Description}
\label{Sec:Dataset}

\begin{center}
    \begin{table}[!t]
\centering
\caption{The real-world deceleration dataset collected from Hyundai electric vehicles}
\begin{tabular}{c|c|c|c}
\hline
\textbf{Vehicle Type} & \textbf{Driver} & \textbf{\begin{tabular}[c]{@{}c@{}}Dataset Size\\ (No. of Profiles)\end{tabular}} & \textbf{Anomaly Ratio} \\ \hline \hline
\textbf{IONIQ5} & \textbf{D1} & 75 & 0.67 \\ \hline
\textbf{IONIQ5} & \textbf{D2} & 25 & 0.60 \\ \hline
\textbf{GV80EV} & \textbf{D1} & 15 & 0.53 \\ \hline
\textbf{GV80EV} & \textbf{D2} & 49 & 0.63 \\ \hline
\end{tabular}
\label{Tab:Data}
\end{table}
\end{center}

\begin{figure*}[!t]
 \centering
 \includegraphics[width=0.85\textwidth]{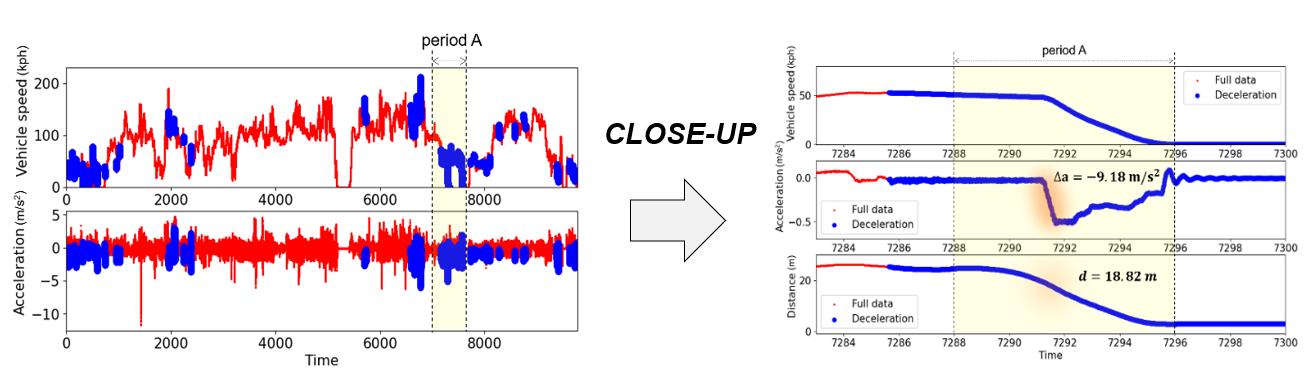}
 \caption{Data preprocessing for deceleration profile generation, where the X-axis is the time, and the Y-axis represents the acceleration, speed, and distance, respectively.}
 \label{Fig:Preprocessing}
\end{figure*}

The real-world deceleration data can more effectively and accurately measure the model's performance. Hence, we collected the real-world deceleration data, which is summarized in Table ~\ref{Tab:Data}. This dataset was collected from August 2023 to May 2024 during actual drives, and data {instances} where the driver's foot was off the accelerator pedal were considered to indicate a lack of intention to accelerate and a desire to reduce speed. Therefore, such instances were acquired for our data. The detailed preprocessing of the collected data involves several steps, starting with synchronizing all the data to a 10$ms$ measurement interval using linear interpolation to match the varying measurement intervals of different vehicles. Subsequently, as shown in Figure ~\ref{Fig:Preprocessing}, the final deceleration profiles were generated based on the following conditions:
\begin{itemize}
    \item \textbf{Step 1.} Extract {data} when the accelerator pedal was not pressed (accelerator pedal stroke < 0.5\%) to generate each deceleration profile.
    \item \textbf{Step 2.} Retain {profiles} when the maximum vehicle speed is above {15 kph} and the brake pedal is pressed more than 2\%.
    \item \textbf{Step 3.} Extract {profiles} when the lateral acceleration is less than {0.07 G}, ensuring that only deceleration data from straight driving is included.
\end{itemize}

The final deceleration datasets consist of two vehicle types (IONIQ5 and GV80EV) and two different drivers (D1 and D2), resulting in four different vehicle type-driver configurations: IONIQ5-D1, IONIQ5-D2, GV80EV-D1, and GV80EV-D2, respectively. The labeling of each dataset was manually conducted by experts in the company. During the labeling process, states showing a similar pattern of the driver pressing the brake pedal, where the longitudinal deceleration values start in negative values then gradually converge to zero, were defined as the driver's braking habits and labeled accordingly.

\section{Preliminaries}
\subsection{Neural Controlled Differential Equations and Adaptive Graph Generation}
{NCDE} have been shown to outperform RNN-based models and Neural Ordinary Differential Equations (NODE) on benchmark datasets over the various MTS domains \cite{kidger2020neural, jhin2023precursor}. In particular, NCDE extends NODE~\cite{chen2018neural} by modeling continuous time series data over all time intervals using differential equations combined with neural networks. To account for temporal dependencies, NCDE creates a continuous path \(\mathbf{X}(t)\) through continuous path interpolation and considers both current and past observations to generate future observations of the time series data. Furthermore, previous research \cite{choi2022graph} combined NCDE with graph operations to process both temporal and spatial information simultaneously. Inspired by this idea, our proposed \fraName~encodes multivariate deceleration data into spatio-temporal features.

Especially, NCDE defines a cubic spline path to view discrete values of a time series as a continuous time series path to estimate continuous dynamics as follows:
\begin{equation}
\label{eq:ncde}
\mathbf{H}(T) = \mathbf{H}(T_0) + \int_{T_0}^{T} f_1(\mathbf{H}(t); \theta_{f_1}) \frac{d\mathbf{X}(t)}{dt} \, dt,
\end{equation}
where \(\mathbf{X}(t)\) is the continuous path of the time series processed by cubic spline interpolation, and \(T\) represents the entire time interval of the time series that is defined in the interval \((T_0, T_n]\).
NCDE integrates the entire path, combining it with the initial value to obtain the final state in a single calculation.
This provides a hidden vector that encodes the temporal information.

In addition, to encode spatial information, we define the following equation to learn the adjacency matrix for the graph structure:
\begin{align}
\mathbf{A} = {softmax(\max(0, \mathbf{E} \cdot \mathbf{E}^\top))},
\end{align}
where \(\mathbf{E} \in \mathbb{R}^{n \times d}\) is the node embedding for each sensor, which yields the normalized graph adjacency matrix \(\mathbf{A}\) after passing through the ReLU and the softmax function. Using \(\mathbf{A}\), we recursively compute the recurrence relation of the Chebyshev polynomial \(\mathbf{C}_k\) within NCDE to efficiently process graph operations~\cite{hammond2011wavelets, kipf2016semi}:
\begin{align}
\mathbf{C}_k = 2 \mathbf{A} \cdot \mathbf{C}_{k-1} - \mathbf{C}_{k-2}, \quad \text{for } k \geq 2 \quad (\mathbf{C}_0 = \mathbf{I}, \, \mathbf{C}_1 = \mathbf{A}),
\end{align}
where \( \mathbf{C}_k \) is the adjacency matrix computed through the Chebyshev polynomial, which defines the graph structure of the deceleration data, representing inter-sensor relations. This process allows us to encode spatial information effectively in the MTS data.

\begin{figure*}[!ht]
 \centering
 \includegraphics[width=0.85\textwidth]{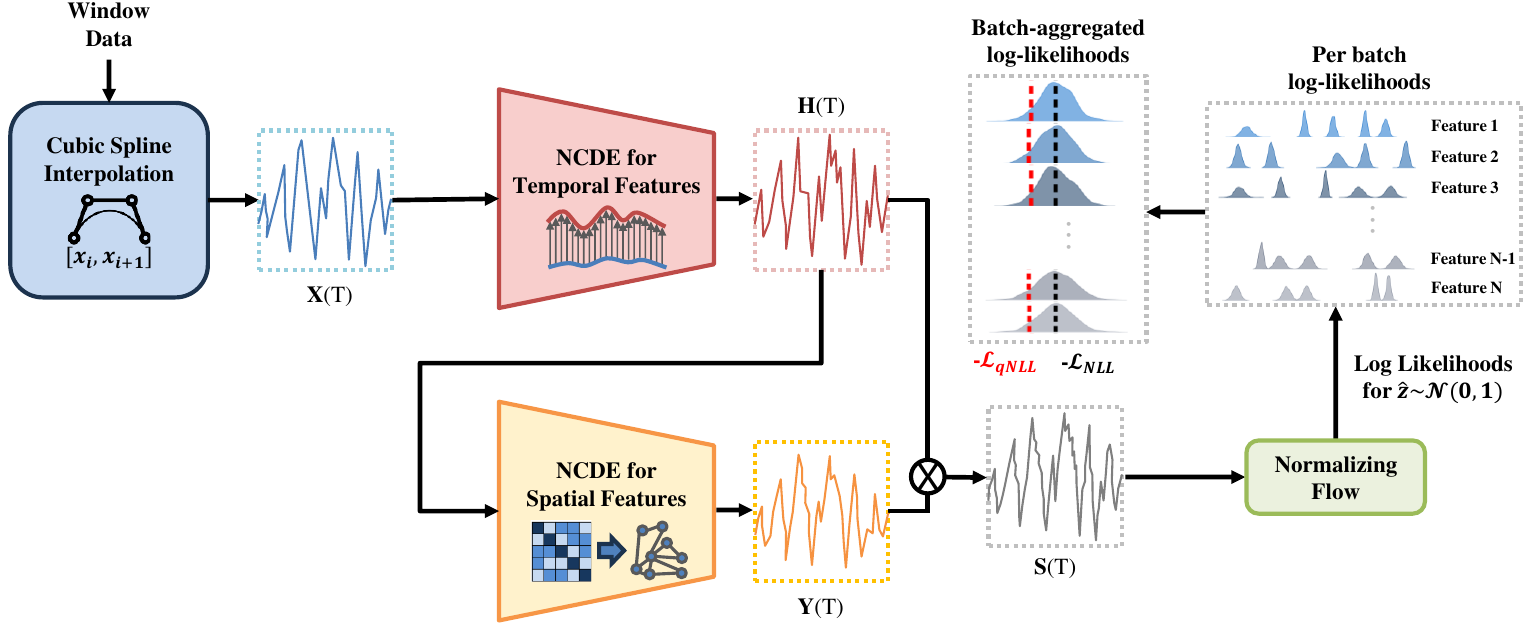}
 \caption{The overall architecture of \fraName. The preprocessed input data \( \mathbf{W}_p^{(i)} \) is first converted into a continuous path \( X(T) \) through cubic spline interpolation. This path passes through two CDE functions to encode spatio-temporal information, resulting in \( H(T) \) and \( Y(T) \), respectively. These are then combined through matrix multiplication to form \( S(T) \), which is used for density estimation in the NF. The log-likelihoods obtained from this process are further processed with a quantile function to produce \( \mathcal{L}_{Q-NLL} \), which is used to detect normal or anomalies based on the threshold \( \tau \).}
 \label{Fig:Framework}
\end{figure*}

\subsection{Normalizing Flow}
NF is a set of reversible transformations \( g_\theta \) that convert the original distribution \( p_{{X}}(\mathbf{x}) \) of the data to a target distribution \( p_{{Z}}(\mathbf{\hat{z}}) \), where \( Z \) follows a standard normal distribution \( \mathcal{N}(0,1) \).
NF gradually transforms the complex distribution of the original data into a simpler distribution of the target distribution, and the density estimation \( p_{{X}}(\mathbf{x}) \) for this transformation is given by the following equation:
\begin{align}
p_{{X}}(\mathbf{x}) = p_{Z}(g_{\theta}(\mathbf{x})) \left| \det \left( \frac{\partial g_{\theta}(\mathbf{x})}{\partial \mathbf{x}} \right) \right|,
\end{align}
where \( g_\theta(\mathbf{x}) \) is the data transformed by the NF \( g_\theta \). The probability density of the original data is calculated through the absolute value of the Jacobian determinant \(\left| \det \left( \frac{\partial g_{\theta}(\mathbf{x})}{\partial \mathbf{x}} \right) \right| \), which signifies density estimation in NF. This process effectively models the complex distribution of data, and in our study, we model the distribution of normal data to identify data that deviates from the normal distribution, performing anomaly detection.
Additionally, the log-density can be expressed as:
\begin{align}
\log p_{{X}}(\mathbf{x}) = \log p_{{Z}}(g_{\theta}(\mathbf{x})) + \log \left| \det \left( \frac{\partial g_{\theta}(\mathbf{x})}{\partial \mathbf{x}} \right) \right|.
\end{align}
Although NF is reversible, our study focuses solely on using forward flow to detect anomalies in driving data by utilizing log-likelihood (LL) estimation without using the inverse.

\section{Our Approach}
\subsection{Problem Definition}
In this section, we define the problem of learning normal driving behavior and identifying anomalies in the deceleration data. The input deceleration data is defined as \( \mathbf{x}_{0:T} = \{\mathbf{x}_0, \mathbf{x}_1, \ldots, \mathbf{x}_T \} \), where \( T \) represents the time interval of the time series. Since the deceleration data is multivariate with \( n \) sensors, the observation \( \mathbf{x}_t \) at each time step is a \( n \)-dimensional vector defined as \( \mathbf{x}_t = [x_t^{(1)}, x_t^{(2)}, \ldots, x_t^{(n)}] \). Additionally, this dataset consists of multiple profiles \( p \), where each profile \( p \) is defined as \( \mathbf{x}_p(t) = [x_p^{(1)}(t), x_p^{(2)}(t), \ldots, x_p^{(n)}(t)] \).
 However, since each \( p \) consists of MTS with varying lengths, we apply a sliding window, resulting in \( \mathbf{W}_p^{(i)} = [ \mathbf{x}_p(i), \mathbf{x}_p(i+1), \cdots, \mathbf{x}_p(i+w-1) ] \). Here, \( w \) is the window size, and \( s \) is the stride. These segments are batched for training, resulting in data with the shape \( \mathbf{W}_p^{(i)} \in \mathbb{R}^{b \times n \times w} \), where \( b \) is the batch size.

\subsection{Overall Architecture}
 Figure~\ref{Fig:Framework} presents the overall architecture of our \fraName. As shown, the initial input data is transformed into a continuous path \(\mathbf{X}(t)\) through cubic spline interpolation. This transformation enables the data to be integrated into the model, continually depending on the observed data rather than changing hidden states. Our model consists of NCDE and AGG modules to capture spatio-temporal information at the same time. When \(\mathbf{X}(t)\) passes through NCDE to extract temporal features, it produces \(\mathbf{H}(T)\), which then passes through NCDE for spatial features to produce \(\mathbf{Y}(T)\). This \(\mathbf{Y}(T)\) is then matrix-multiplied with \(\mathbf{H}(T)\) to create \(\mathbf{S}(T)\), which contains the spatio-temporal information. 
 
 The latent vectors produced through this process are transformed by NF to map the original data distribution closer to a normal distribution. The resulting {LL} indicates that the closer the model's latent representation is to a simple {distribution,} such as the normal distribution, the higher the value {is} in the positive direction.
 However, typical deep learning models optimize by minimizing the objective function, so we train the NF by minimizing the negative log-likelihood (NLL), ensuring convergence in the intended direction.
 Additionally, we apply the quantile function to LL to better separate normal data near the boundary of the normal distribution from anomalies.

\subsection{Spatio-Temporal Information Encoding}
We use two CDE functions, \( f_1 \) and \( f_2 \), to simultaneously encode temporal and spatial information. The process for encoding temporal information, \( \mathbf{H}(T) \), is the same as in Eq.~(\ref{eq:ncde}), while the process for encoding spatial information, \( \mathbf{Y}(T) \), is defined as follows:
\begin{align}
\mathbf{Y}(T) = \mathbf{Y}(T_0) + \int_{T_0}^{T} f_2(\mathbf{Y}(t); \theta_{f_2}) \frac{d\mathbf{H}(t)}{dt} \, dt,
\textbf{} \end{align}
where the CDE function \( f \) consists of fully connected layers, and each row of the encoded temporal information \( \mathbf{H}(t) \) is taken as an individual input. The CDE function \( f_1 \) encodes temporal information, and the CDE function \( f_2 \) encodes spatial information. The encoded \( \mathbf{H}(T) \) and \( \mathbf{Y}(T) \) from the two CDE functions are then multiplied together to produce \( \mathbf{S}(T) \). This allows the parameters of the NCDE to be updated using \( \mathbf{S}(t+\Delta t) \) as shown in the following equations:
\begin{align}
\mathbf{H}(t + \Delta t) = \mathbf{H}(t) + f_1(\mathbf{H}(t); \theta_{f_1}) \cdot \Delta \mathbf{X}(t),
\end{align}
\begin{align}
\mathbf{Y}(t + \Delta t) = \mathbf{Y}(t) + f_2(\mathbf{Y}(t); \theta_{f_2}) \cdot \Delta \mathbf{H}(t),
\end{align}
\begin{align}
\mathbf{S}(t + \Delta t) = \mathbf{Y}(t + \Delta t) \cdot \mathbf{H}(t + \Delta t).
\end{align}
By encoding spatio-temporal information simultaneously and passing it to the NF, we can model the continuous changes in deceleration data more naturally. Additionally, we structure the \( n \) sensors as a graph for learning.

\subsection{Density Estimation}
We use the hidden vectors \( \mathbf{S}(t) \) from NCDE as input to the NF to learn the distributions of MTS containing spatio-temporal information to approximate these distributions closer to the target. To achieve this, we define the target distribution as \( P_{{Z}}(\hat{z}) = \mathcal{N}(\hat{z} | \mu, \mathbf{\Sigma}) \), where \( \mathcal{N} \) is a multivariate normal distribution for MTS, with \( \mu \) as the mean vector and \( \mathbf{\Sigma} \) as the covariance matrix. Thus, the LL of this multivariate normal distribution can be defined as \( \log P_{{Z}}(\hat{z}) = \log \mathcal{N}(\hat{z} | \mu, \mathbf{\Sigma}) \). To transform the original distribution \( p_{{X}}(\mathbf{x}) \) to the target distribution, the density transformation process using the hidden vectors \( S(t) \) as input is as follows:
\begin{equation}
\log P_{X}(\mathbf{S}) = \log P_{Z}(g_\theta(\mathbf{S})) + \log \left| \det \left( \frac{\partial g_\theta(\mathbf{S})}{\partial \mathbf{S}} \right) \right|,
\end{equation}
where \( \log P_{{Z}}(g_\theta(S)) \) is the LL of the transformed latent variable \( \hat{z} \). We calculate these LL for each sensor in the multivariate deceleration data.

\subsection{Quantile-based Maximum Likelihood}
  Since some normal data close to the boundary of the anomalous pattern distribution may have low LL, we use a quantile function to maximize the relatively low LL of normal data to better separate normal and anomalous data. It is important to note that we use the quantile function as an additional computation on the LL score, not for quantile forecasting{~\cite{wen2017multi, kan2022multivariate}}. The LL is optimized in the maximizing direction, but we negate it to create an NLL objective for minimization. We first compute the LL from the NF, apply the quantile function, and then {make it negative} to obtain the \(\mathcal{L}_{Q-NLL}\), as follows:
\begin{align}
\mathcal{L}_{Q-NLL}(Z;\theta) = -\psi \left( \log P_{X}(\mathbf{S}), q\right),
\end{align}
where \( \psi \) denotes the quantile function, which calculates the q-quantile value of the LL to update the model parameters. A smaller \( q \) increases the LL of normal data near the boundary, assuming that the LL of normal data is higher than that of anomalous data. We distinguish between normal and anomalies based on the threshold \( \tau \). Specifically, data points are determined to be normal if \( \mathcal{L}_{Q-NLL} \leq \tau \), and anomalous if \( \mathcal{L}_{Q-NLL} > \tau \).

\section{Experiments}
\subsection{Datasets and Baselines}
\textbf{Datasets.}
We utilized four deceleration datasets collected from two vehicle types (IONIQ5 and GV80EV) and two different drivers (D1 and D2) as introduced in Section ~\ref{Sec:Dataset}: IONIQ5-D1, IONIQ5-D2, GV80EV-D1, and GV80EV-D2.
We designed two experimental configurations to evaluate data cleaning performance and generalization ability. First, in Section~\ref{Sec:Hyperparameter}, we conducted four separate experiments for each deceleration dataset to assess GDFlow's data cleaning performance and hyperparameter sensitivity. Following unsupervised anomaly detection conventions, training data consisted solely of normal data. Second, in Section~\ref{Sec:Generalization}, we used a holdout approach to evaluate generalization. One deceleration dataset was held out as test data while the others were used for training, assessing the model's performance on unseen vehicle-driver combinations. 

Additionally, we quantitatively compared GDFlow's anomaly detection efficacy against standard anomaly detection benchmark datasets. For this comparison, we used the SMD~\cite{su2019robust}, MSL~\cite{hundman2018detecting}, and SMAP~\cite{hundman2018detecting} datasets. These datasets exhibit significant distribution shifts similar to real-world data.
Details of the benchmark datasets are described in Appendix~\ref{sec:bench}.

\noindent \textbf{Baselines.}
We compared a range of selected {SOTA} anomaly detection algorithms against our proposed GDFlow. For comparing anomaly detection performance, we used five baselines: THOC~\cite{shen2020timeseries}, AT~\cite{xu2021anomaly}, M2N2~\cite{kim2024model}, GANF~\cite{dai2022graph}, and MTGFLOW~\cite{zhou2023detecting}, respectively. Note that for a comprehensive comparison, we included traditional anomaly detection approaches such as reconstruction and association-based methods, as well as newly proposed NF-based approaches. From these, the two top-performing anomaly detector baselines on the anomaly detection benchmarks were then utilized for performance comparison on our deceleration datasets.

To demonstrate the superiority of our anomaly detection performance over other methods, we also employed DBSCAN-DTW, SAE-DBSCAN, GAE-DBSCAN, and DCEC~\cite{aytekin2018clustering} for comparison. DBSCAN-DTW uses DTW instead of Euclidean distance for better time series modeling. SAE-DBSCAN and GAE-DBSCAN are autoencoder-enhanced DBSCAN algorithms using sparse autoencoder and GRU-autoencoders, respectively.

\subsection{Evaluation Metrics}
To evaluate the effectiveness of \fraName~against competing models, we utilize the F1-score, {Area Under the Receiver Operating Characteristics curve (AUROC), and Area Under the Precision-Recall Curve (AUPRC)} metrics. After analyzing the test data statistics, the best threshold is determined for the F1-score to comprehensively evaluate precision and recall through their harmonic mean. Since real-world anomalies typically span multiple timestamps, we apply a point-adjusted assessment protocol, considering the identification of any part of an anomaly segment as a correct detection~\cite{su2019robust,xu2018unsupervised}. 

{Additionally}, we report AUROC and AUPRC over test data anomaly scores, providing an overall summary of the anomaly detector's performance across all possible thresholds. AUROC evaluates performance across a range of decision thresholds, reducing sensitivity to a specific threshold choice. AUPRC is particularly suited for imbalanced classification scenarios, offering insights into the model's precision and recall balance.
{Given that \fraName~is an NF-based anomaly detector, we selected GANF and MTGFLOW as our primary competing baselines. In experiments on benchmark datasets, we used only AUROC for comparison, following previous research works~\cite{dai2022graph,zhou2023detecting}.}
\begin{table*}[!t]
\centering
\caption{Anomaly detection performance and hyperparameter sensitivity on individual deceleration datasets. The best performance is highlighted in bold, and the second-best performance is underlined.}
\small
\begin{tabular}{clccccc}
\hline
\multicolumn{2}{c|}{\textbf{Dataset}} &
  \multicolumn{2}{c|}{\textbf{Method}} &
  \textbf{F1-PA} &
  \textbf{AUROC} &
  \textbf{AURPC} \\ \hline \hline
\multicolumn{2}{c|}{\multirow{6}{*}{\textbf{\begin{tabular}[c]{@{}c@{}}Train:\\ IONIQ-D1 with normal 50\%\\ \\ Test:\\ IONIQ-D1 with anomaly 100\%\\ + normal 50\%\end{tabular}}}} &
  \multicolumn{1}{c|}{\multirow{4}{*}{\textbf{Clustering}}} &
  \multicolumn{1}{c|}{\textbf{DBSCAN-DTW}} &
  0.8540±0.1160 &
  0.5435±0.0521 &
  \textbf{0.9128±0.0090} \\
\multicolumn{2}{c|}{} &
  \multicolumn{1}{c|}{} &
  \multicolumn{1}{c|}{\textbf{SAE-DBSCAN}} &
  0.1004±0.1460 &
  0.5125±0.0197 &
  0.7391±0.2134 \\
\multicolumn{2}{c|}{} &
  \multicolumn{1}{c|}{} &
  \multicolumn{1}{c|}{\textbf{GAE-DBSCAN}} &
  0.0530±0.1302 &
  0.5153±0.0377 &
  0.9138±0.0049 \\
\multicolumn{2}{c|}{} &
  \multicolumn{1}{c|}{} &
  \multicolumn{1}{c|}{\textbf{DCEC}} &
  0.9043±0.0028 &
  0.5672±0.0829 &
  0.8561±0.0391 \\ \cline{3-7} 
\multicolumn{2}{c|}{} &
  \multicolumn{1}{c|}{\multirow{2}{*}{\textbf{\begin{tabular}[c]{@{}c@{}}Anomaly\\ Detection\end{tabular}}}} &
  \multicolumn{1}{c|}{\textbf{M2N2}} &
  \textbf{0.9970±0.0023} &
  0.4871±0.0553 &
  0.7962±0.0367 \\
\multicolumn{2}{c|}{} &
  \multicolumn{1}{c|}{} &
  \multicolumn{1}{c|}{\textbf{MTGFLOW}} &
  0.9036±0.0015 &
  \textbf{0.6500±0.0639} &
  \underline{0.9076±0.0244} \\ \hline
\multicolumn{4}{c|}{\textbf{GDFlow (Ours)}} &
  \underline{0.9056±0.0040} &
  \underline{0.6398±0.0257} &
  0.9041±0.0091 \\ \hline
\multicolumn{2}{c|}{\multirow{6}{*}{\textbf{\begin{tabular}[c]{@{}c@{}}Train:\\ IONIQ5-D2 with normal 50\%\\ \\ Test:\\ IONIQ5-D2 with anomaly 100\%\\ + normal 50\%\end{tabular}}}} &
  \multicolumn{1}{c|}{\multirow{4}{*}{\textbf{Clustering}}} &
  \multicolumn{1}{c|}{\textbf{DBSCAN-DTW}} &
  0.9394±0.0000 &
  0.5000±0.0000 &
  0.9429±0.0000 \\
\multicolumn{2}{c|}{} &
  \multicolumn{1}{c|}{} &
  \multicolumn{1}{c|}{\textbf{SAE-DBSCAN}} &
  0.7454±0.2716 &
  0.6684±0.1373 &
  \underline{0.9617±0.0155} \\
\multicolumn{2}{c|}{} &
  \multicolumn{1}{c|}{} &
  \multicolumn{1}{c|}{\textbf{GAE-DBSCAN}} &
  0.3707±0.4342 &
  0.5596±0.1042 &
  0.9489±0.0115 \\
\multicolumn{2}{c|}{} &
  \multicolumn{1}{c|}{} &
  \multicolumn{1}{c|}{\textbf{DCEC}} &
  0.9453±0.0090 &
  0.5277±0.1362 &
  0.8915±0.0442 \\ \cline{3-7} 
\multicolumn{2}{c|}{} &
  \multicolumn{1}{c|}{\multirow{2}{*}{\textbf{\begin{tabular}[c]{@{}c@{}}Anomaly\\ Detection\end{tabular}}}} &
  \multicolumn{1}{c|}{\textbf{M2N2}} &
  \textbf{0.9931±0.0040} &
  0.4529±0.1268 &
  0.8068±0.0557 \\
\multicolumn{2}{c|}{} &
  \multicolumn{1}{c|}{} &
  \multicolumn{1}{c|}{\textbf{MTGFLOW}} &
  0.9599±0.0097 &
  \underline{0.7272±0.0631} &
  0.9569±0.0196 \\ \hline
\multicolumn{4}{c|}{\textbf{GDFlow (Ours)}} &
  \underline{0.9680±0.0118} &
  \textbf{0.7430±0.0436} &
  \textbf{0.9629±0.0096} \\ \hline
\multicolumn{2}{c|}{\multirow{6}{*}{\textbf{\begin{tabular}[c]{@{}c@{}}Train:\\ GV80EV-D1 with normal 50\%\\ \\ Test:\\ GV80EV-D1 with anomaly 100\%\\ + normal 50\%\end{tabular}}}} &
  \multicolumn{1}{c|}{\multirow{4}{*}{\textbf{Clustering}}} &
  \multicolumn{1}{c|}{\textbf{DBSCAN-DTW}} &
  0.9873±0.0034 &
  0.5516±0.1192 &
  0.9875±0.0033 \\
\multicolumn{2}{c|}{} &
  \multicolumn{1}{c|}{} &
  \multicolumn{1}{c|}{\textbf{SAE-DBSCAN}} &
  0.2093±0.3328 &
  0.5214±0.1165 &
  0.9790±0.0217 \\
\multicolumn{2}{c|}{} &
  \multicolumn{1}{c|}{} &
  \multicolumn{1}{c|}{\textbf{GAE-DBSCAN}} &
  0.1396±0.3503 &
  0.5233±0.0586 &
  0.9867±0.0015 \\
\multicolumn{2}{c|}{} &
  \multicolumn{1}{c|}{} &
  \multicolumn{1}{c|}{\textbf{DCEC}} &
  0.9886±0.0026 &
  0.6193±0.2413 &
  0.9743±0.0192 \\ \cline{3-7} 
\multicolumn{2}{c|}{} &
  \multicolumn{1}{c|}{\multirow{2}{*}{\textbf{\begin{tabular}[c]{@{}c@{}}Anomaly\\ Detection\end{tabular}}}} &
  \multicolumn{1}{c|}{\textbf{M2N2}} &
  \textbf{0.9978±0.0034} &
  0.1457±0.0347 &
  0.8940±0.0134 \\
\multicolumn{2}{c|}{} &
  \multicolumn{1}{c|}{} &
  \multicolumn{1}{c|}{\textbf{MTGFLOW}} &
  \underline{0.9942±0.0015} &
  \underline{0.9725±0.0275} &
  \underline{0.9992±0.0008} \\ \hline
\multicolumn{4}{c|}{\textbf{GDFlow (Ours)}} &
  0.9937±0.0005 &
  \textbf{0.9817±0.0031} &
  \textbf{0.9995±0.0001} \\ \hline
\multicolumn{2}{c|}{\multirow{6}{*}{\textbf{\begin{tabular}[c]{@{}c@{}}Train:\\ GV80EV-D2 with normal 50\%\\ \\ Test:\\ GV80EV-D2 with anomaly 100\%\\ + normal 50\%\end{tabular}}}} &
  \multicolumn{1}{c|}{\multirow{4}{*}{\textbf{Clustering}}} &
  \multicolumn{1}{c|}{\textbf{DBSCAN-DTW}} &
  0.9333±0.0140 &
  0.5166±0.0276 &
  \underline{0.9427±0.0015} \\
\multicolumn{2}{c|}{} &
  \multicolumn{1}{c|}{} &
  \multicolumn{1}{c|}{\textbf{SAE-DBSCAN}} &
  0.4997±0.2386 &
  0.6081±0.0990 &
  0.9346±0.0240 \\
\multicolumn{2}{c|}{} &
  \multicolumn{1}{c|}{} &
  \multicolumn{1}{c|}{\textbf{GAE-DBSCAN}} &
  0.1988±0.2748 &
  0.5501±0.0692 &
  \textbf{0.9440±0.0061} \\
\multicolumn{2}{c|}{} &
  \multicolumn{1}{c|}{} &
  \multicolumn{1}{c|}{\textbf{DCEC}} &
  \underline{0.9449±0.0059} &
  0.4462±0.0698 &
  0.8674±0.0267 \\ \cline{3-7} 
\multicolumn{2}{c|}{} &
  \multicolumn{1}{c|}{\multirow{2}{*}{\textbf{\begin{tabular}[c]{@{}c@{}}Anomaly\\ Detection\end{tabular}}}} &
  \multicolumn{1}{c|}{\textbf{M2N2}} &
  \textbf{0.9976±0.0007} &
  0.4190±0.0628 &
  0.8345±0.0278 \\
\multicolumn{2}{c|}{} &
  \multicolumn{1}{c|}{} &
  \multicolumn{1}{c|}{\textbf{MTGFLOW}} &
  0.9394±0.0029 &
  \underline{0.6413±0.0883} &
  0.9320±0.0226 \\ \hline \hline
\multicolumn{4}{c|}{\textbf{GDFlow (Ours)}} &
  0.9431±0.0069 &
  \textbf{0.6594±0.0599} &
  0.9347±0.0125 \\ \hline
\end{tabular}
\label{Tab:sensitivity}
\end{table*}
\begin{table*}[!t]
\centering
\caption{Anomaly detection performance on cross-deceleration datasets. The best performance is highlighted in bold, and the second-best performance is underlined.}
\small
\begin{tabular}{clcc|ccc}
\hline
\multicolumn{2}{c|}{\textbf{Dataset}} &
  \multicolumn{2}{c|}{\textbf{Method}} &
  \textbf{F1-PA} &
  \textbf{AUROC} &
  \textbf{AURPC} \\ \hline \hline
\multicolumn{2}{c|}{\multirow{6}{*}{\textbf{\begin{tabular}[c]{@{}c@{}}Train:\\ IONIQ-D2,\\ GV80EV-D1\&D2\\ \\ Test:\\ IONIQ-D1\end{tabular}}}} &
  \multicolumn{1}{c|}{\multirow{4}{*}{\textbf{Clustering}}} &
  \textbf{DBSCAN-DTW} &
  0.7473 &
  0.4805 &
  \underline{0.8022} \\
\multicolumn{2}{c|}{} &
  \multicolumn{1}{c|}{} &
  \textbf{SAE-DBSCAN} &
  0.0128 &
  0.4891 &
  0.4611 \\
\multicolumn{2}{c|}{} &
  \multicolumn{1}{c|}{} &
  \textbf{GAE-DBSCAN} &
  0.0130 &
  0.5033 &
  \textbf{0.8182} \\
\multicolumn{2}{c|}{} &
  \multicolumn{1}{c|}{} &
  \textbf{DCEC} &
  \underline{0.7838} &
  0.5014 &
  0.6628 \\ \cline{3-7} 
\multicolumn{2}{c|}{} &
  \multicolumn{1}{c|}{\multirow{2}{*}{\textbf{\begin{tabular}[c]{@{}c@{}}Anomaly\\ Detection\end{tabular}}}} &
  \textbf{M2N2} &
  \textbf{0.9807} &
  0.4575 &
  0.6337 \\
\multicolumn{2}{c|}{} &
  \multicolumn{1}{c|}{} &
  \textbf{MTGFLOW} &
  0.7795 &
  \underline{0.5783} &
  0.7061 \\ \hline
\multicolumn{4}{c|}{\textbf{GDFlow (ours)}} &
  0.7795 &
  \textbf{0.6038} &
  0.7063 \\ \hline
\multicolumn{2}{c|}{\multirow{6}{*}{\textbf{\begin{tabular}[c]{@{}c@{}}Train:\\ IONIQ5-D1,\\ GV80EV-D1\&D2\\ \\ Test:\\ IONIQ5-D2\end{tabular}}}} &
  \multicolumn{1}{c|}{\multirow{4}{*}{\textbf{Clustering}}} &
  \textbf{DBSCAN-DTW} &
  0.8228 &
  0.5000 &
  0.8495 \\
\multicolumn{2}{c|}{} &
  \multicolumn{1}{c|}{} &
  \textbf{SAE-DBSCAN} &
  0.7289 &
  \underline{0.7867} &
  \textbf{0.9358} \\
\multicolumn{2}{c|}{} &
  \multicolumn{1}{c|}{} &
  \textbf{GAE-DBSCAN} &
  0.7034 &
  0.7204 &
  0.8905 \\
\multicolumn{2}{c|}{} &
  \multicolumn{1}{c|}{} &
  \textbf{DCEC} &
  0.8527 &
  0.3312 &
  0.6288 \\ \cline{3-7} 
\multicolumn{2}{c|}{} &
  \multicolumn{1}{c|}{\multirow{2}{*}{\textbf{\begin{tabular}[c]{@{}c@{}}Anomaly\\ Detection\end{tabular}}}} &
  \textbf{M2N2} &
  \textbf{0.9942} &
  0.3982 &
  0.6456 \\
\multicolumn{2}{c|}{} &
  \multicolumn{1}{c|}{} &
  \textbf{MTGFLOW} &
  0.8519 &
  0.7725 &
  0.9067 \\ \hline
\multicolumn{4}{c|}{\textbf{GDFlow (ours)}} &
  \underline{0.9163} &
  \textbf{0.8056} &
  \underline{0.9188} \\ \hline
\multicolumn{2}{c|}{\multirow{6}{*}{\textbf{\begin{tabular}[c]{@{}c@{}}Train:\\ IONIQ5-D1\&D2,\\ GV80EV-D2\\ \\ Test:\\ GV80EV-D1\end{tabular}}}} &
  \multicolumn{1}{c|}{\multirow{4}{*}{\textbf{Clustering}}} &
  \textbf{DBSCAN-DTW} &
  0.9701 &
  \underline{0.5000} &
  \underline{0.9710} \\
\multicolumn{2}{c|}{} &
  \multicolumn{1}{c|}{} &
  \textbf{SAE-DBSCAN} &
  0.0130 &
  0.2134 &
  0.5509 \\
\multicolumn{2}{c|}{} &
  \multicolumn{1}{c|}{} &
  \textbf{GAE-DBSCAN} &
  0.0133 &
  0.3834 &
  0.6284 \\
\multicolumn{2}{c|}{} &
  \multicolumn{1}{c|}{} &
  \textbf{DCEC} &
  \underline{0.9850} &
  0.1149 &
  0.8219 \\ \cline{3-7} 
\multicolumn{2}{c|}{} &
  \multicolumn{1}{c|}{\multirow{2}{*}{\textbf{\begin{tabular}[c]{@{}c@{}}Anomaly\\ Detection\end{tabular}}}} &
  \textbf{M2N2} &
  \textbf{0.9954} &
  0.1393 &
  0.8057 \\
\multicolumn{2}{c|}{} &
  \multicolumn{1}{c|}{} &
  \textbf{MTGFLOW} &
  0.9754 &
  0.2163 &
  0.8599 \\ \hline
\multicolumn{4}{c|}{\textbf{GDFlow (ours)}} &
  \textbf{0.9954} &
  \textbf{0.9550} &
  \textbf{0.9956} \\ \hline
\multicolumn{2}{c|}{\multirow{6}{*}{\textbf{\begin{tabular}[c]{@{}c@{}}Train:\\ IONIQ5-D1\&D2,\\ GV80EV-D1\\ \\ Test:\\ GV80EV-D2\end{tabular}}}} &
  \multicolumn{1}{c|}{\multirow{4}{*}{\textbf{Clustering}}} &
  \textbf{DBSCAN-DTW} &
  0.8059 &
  0.4695 &
  0.8583 \\
\multicolumn{2}{c|}{} &
  \multicolumn{1}{c|}{} &
  \textbf{SAE-DBSCAN} &
  0.4132 &
  \textbf{0.6302} &
  \textbf{0.9109} \\
\multicolumn{2}{c|}{} &
  \multicolumn{1}{c|}{} &
  \textbf{GAE-DBSCAN} &
  0.3866 &
  \underline{0.6198} &
  \underline{0.9084} \\
\multicolumn{2}{c|}{} &
  \multicolumn{1}{c|}{} &
  \textbf{DCEC} &
  0.8557 &
  0.4342 &
  0.7259 \\ \cline{3-7} 
\multicolumn{2}{c|}{} &
  \multicolumn{1}{c|}{\multirow{2}{*}{\textbf{\begin{tabular}[c]{@{}c@{}}Anomaly\\ Detection\end{tabular}}}} &
  \textbf{M2N2} &
  \textbf{0.9773} &
  0.4067 &
  0.7220 \\
\multicolumn{2}{c|}{} &
  \multicolumn{1}{c|}{} &
  \textbf{MTGFLOW} &
  0.8631 &
  0.5594 &
  0.7880 \\ \hline \hline
\multicolumn{4}{c|}{\textbf{GDFlow (Ours)}} &
  \underline{0.8750} &
  0.6083 &
  0.8151 \\ \hline
\end{tabular}
\label{Tab:generality}
\end{table*}
\begin{table*}[!t]
\centering
\caption{AUROC performance and ablation study on four real-world benchmark datasets. The best performance is highlighted in bold, and the second-best performance is underlined. Note that `w/o' denotes `without', and `\(\textbf{Q}\)' denotes the `quantile function'.}
\small
\begin{tabular}{c|ccccc|c|ccc}
\hline
\textbf{Dataset} & \textbf{THOC} & \multicolumn{1}{l}{\textbf{AT}} & \textbf{M2N2} & \textbf{GANF} & \textbf{MTGFLOW} & \textbf{GDFlow (Ours)} & \textbf{w/o NCDE} & \textbf{w/o Q} & \textbf{w/o NCDE \& Q} \\ \hline \hline
\textbf{SMD (M-1-4)} & 0.869 & 0.479 & 0.845       & 0.785       & \underline{0.881} & \textbf{0.932} & 0.528 & 0.500 & 0.500 \\
\textbf{SMD (M-2-1)} & 0.668 & 0.498 & \underline{0.764} & 0.692       & 0.754       & \textbf{0.788} & 0.503 & 0.500 & 0.500 \\
\textbf{MSL (P-15)}  & 0.332 & 0.568 & \underline{0.801} & 0.482       & 0.482       & \textbf{0.975} & 0.918 & 0.493 & 0.488 \\
\textbf{SMAP (T-3)}  & 0.591 & 0.490 & 0.617       & \underline{0.780} & 0.775       & \textbf{0.801} & 0.781 & 0.783 & 0.760 \\ \hline
\end{tabular}
\label{Tab:benchmarks}
\end{table*}
\subsection{Evaluation with Deceleration Datasets}
All experiments are conducted with a fixed seed, and {the} hyperparameters and implementation details for each model are presented in Appendix \ref{sec:a1}, \ref{sec:a2}, and \ref{sec:a3}.
\subsubsection{Data Cleaning Performance and Hyperparameter Sensitivity}
\label{Sec:Hyperparameter}
In this section, we present the evaluation results in Table~\ref{Tab:sensitivity}, showing the performance gain of the data cleaning process and hyperparameter sensitivity.
We measure the mean and standard deviation under the various hyperparameter configurations to optimize the experiments.

Across the four deceleration datasets, the M2N2 model achieved the best performance in terms of F1-PA; however, its {mean} AUROC performance across the datasets did not exceed 0.5. In contrast, our proposed model achieved the second-best performance on IONIQ5-D1 and the highest AUROC performance on the other three datasets. Regarding AURPC, DBSCAN-DTW, and GAE-DBSCAN achieved the best performance on IONIQ5-D1 and GV80EV-D2, respectively. However, the difference from our model was only about 0.01. Our model achieved the highest performance on IONIQ5-D2 and GV80EV-D1, particularly excelling on GV80EV-D1 with an AUPRC of 0.9995.

The standard deviations (values following ±) were generally higher for clustering models. 
Specifically, GAE-DBSCAN showed a high standard deviation of 0.4342 for F1-PA on the IONIQ5-D2 dataset, indicating significant hyperparameter sensitivity even with the best threshold for F1-score in each experiment. 
Additionally, DBSCAN-DTW, used in existing ACC, also exhibited generally high standard deviations.
Moreover, for the IONIQ5-D2 dataset, the mean AUROC was 0.5 with a standard deviation of 0, indicating the worst-case scenario where DBSCAN-DTW consistently assigned all data as normal.
Nonetheless, this not only implies high hyperparameter sensitivity for DBSCAN-DTW, but also suggests significant variability in hyperparameter ranges even within datasets of the same domain.

In contrast, our model generally maintained low standard deviation values. While M2N2, MTGFLOW, GV80EV-D1, and DCEC showed lower standard deviations than our model for F1-PA, our model maintained lower standard deviations with AUROC and AURPC. Since the F1-score measures performance based on a single best threshold, other models may yield higher scores.
{However, since it is difficult to select the optimal threshold for every dataset in a realistic scenario, AUROC, which measures the overall performance across various thresholds, and AUPRC, which considers precision and recall across all ranges, are much better metrics to evaluate hyperparameter sensitivity.}
Therefore, we emphasize that AUROC and AURPC are most suitable, realistic, and practical for measuring hyperparameter sensitivity in MTS anomaly detection. Hence, the lower standard deviation of our model in these metrics suggests that our model has the lowest hyperparameter sensitivity.

\subsubsection{Generalization Ability}
\label{Sec:Generalization}
Table \ref{Tab:generality} shows the F1-PA, AUROC, and AURPC performance of six baselines and \fraName~on four different dataset configurations. Our model achieves the highest AUROC on three datasets and shows only a marginal difference {(0.0219) with upper bound} on the fourth dataset. In the third dataset, our model outperforms all others in every aspect, particularly in AUROC, where the performance gap between our model (0.9550) and the second-best model (0.5) is notably significant.
Despite its use in the existing ACC, DBSCAN-DTW shows the worst AUROC results of 0.5 in the second and third datasets, indicating its inability to generalize across different vehicle-driver combinations. This limitation arises from its requirement for meticulous hyperparameter tuning in each new scenario, rendering it impractical for ADAS.
In terms of F1-PA and AUPRC, our model consistently demonstrates strong performance across most datasets, showcasing stable and generalized performance even on unseen datasets.

\subsection{Evaluation with Benchmark Datasets}
To demonstrate that our model's anomaly detection capabilities are effective not only for deceleration datasets but also in other real-world scenarios, we conducted experiments on four MTS benchmark datasets. Table \ref{Tab:benchmarks} reports the AUROC performance of our model and five {SOTA} anomaly detection baselines on these four benchmark datasets. GANF and MTGFLOW, which are also NF-based models, show high performance, particularly MTGFLOW on SMD (Machine-1-4) and GANF on SMAP, indicating that NF-based models are highly effective for MTS anomaly detection in real-world scenarios. 

On the other hand, our model outperforms all baselines across the four benchmark datasets, demonstrating its superiority over existing NF models and highlighting its applicability and reliability in real-world scenarios.

\subsection{Ablation Study}
To verify the performance improvement resulting from the incorporation of NCDE and the quantile function, we conducted an ablation study with three scenarios: \fraName~ without NCDE (w/o NCDE), \fraName~ without the Quantile function (w/o Q), and \fraName~ without both (w/o NCDE \& Q). Without NCDE, a vanilla RNN was used to handle the temporal information modeling. Table \ref{Tab:benchmarks} presents the ablation study results on benchmark datasets, highlighting the impact of NCDE and the quantile function. In our experiments, the SMD dataset showed the worst AUROC of 0.5 when the quantile function was not used, detecting every profile as normal. Performance degradation was also seen in the MSL and SMAP datasets, especially when neither NCDE nor the quantile function was used. These results indicate that the quantile function is crucial for training stability. {A detailed analysis of NCDE is further presented in Appendix~\ref{sec:mem}.}

\section{Conclusion}
In this work, we propose GDFlow to effectively detect the anomaly patterns in the vehicle braking system as a part of Adaptive Cruise Control (ACC) in Advanced Driver Assistance Systems (ADAS). We aim to tackle the inherent challenges of previous clustering algorithms, which struggle with sensitivity to hyperparameters and difficulty in handling multivariate deceleration profiles. Our approach accurately learns the distribution of normal driving patterns, and simultaneously captures spatio-temporal information from sensor data, allowing for more precise modeling of continuous changes in driving behavior. Furthermore, we enhanced the anomaly detection capabilities of our model by incorporating a novel quantile-based maximum log-likelihood objective, which enables better identification of normal driving data. Our extensive experimental results with the dataset collected from real-world operating vehicles, as well as popular benchmarks, demonstrated that \fraName~outperforms {SOTA} baselines, achieving superior detection performance. Additionally, \fraName~shows significant improvements in hyperparameter sensitivity and potential for generalization to unseen vehicle-driver configurations. With increased efficiency in terms of inference time, it plans to be deployed in March 2026. We strongly believe that \fraName~will pave the way for more reliable and efficient anomaly detection in ADAS, ultimately enhancing driving safety and user experience.

\section*{Acknowledgements}
This research was generously supported by Hyundai Motors Group.

\bibliographystyle{abbrv}
\bibliography{references}

\vspace{3.5mm}

\appendix{\huge \textbf{Appendix}}

\section{Experimental Details for Deceleration Datasets}
The experiments were conducted on a system with the following specifications: Python 3.10.13, NumPy 1.26.2, SciPy 1.11.4, Matplotlib 3.8.2, PyTorch 1.12.1, CUDA 11.3, torchdiffeq 0.2.4, Ubuntu 20.04.5 LTS, AMD EPYC 7352 24-core processor, NVIDIA driver 470.141.03, and NVIDIA RTX A5000 24GB.

We use the AdamW optimizer with a weight decay of 5e-4. The batch size is set to 256, and the window size is set as the shortest profile length in the deceleration data with stride 1. We use 80\% of the entire data as a train split and train the model for 10 epochs. During training, we save the model at each epoch if it achieves the best F1-score, and the saved best model is used for testing.
\subsection{Experimental Setup for Data Cleaning Performance}
\label{sec:a1}
We explored hyperparameter sensitivity using the following combinations of hyperparameters for each model:

\begin{itemize}
\item \textbf{DBSCAN-DTW (using sklearn):} We vary the values of the hyperparameters min samples \(\in\) \{2, 3, 4\} and $\epsilon$ \(\in\) \{0.3, 0.5, 0.7, 1.0, 1.5, 2.0, 3.0\}, respectively.

\item  \textbf{Sparse Autoencoder-DBSCAN (SAE-DBSCAN):} We vary the values of the hyperparameters min samples \(\in\) \{2, 3, 4\} and  $\epsilon$ \(\in\) \{0.3, 0.5, 0.7, 1.0, 1.5, 2.0, 3.0\}, respectively.

\item \textbf{GRU-Autoencoder-DBSCAN (GAE-DBSCAN):} We vary the values of the hyperparameters min samples \(\in\) \{2, 3, 4\} and $\epsilon$ \(\in\) \{0.3, 0.5, 0.7, 1.0, 1.5, 2.0, 3.0\}, respectively.

\item \textbf{Deep Convolutional Embedded Clustering (DCEC):} We vary the values of the hyperparameters \( k \in \{2, 3, 4\} \), kernel sizes \(\in\) \{(1, 1, 1), (3, 3, 1), (5, 5, 3), (7, 7, 5)\}, stride sizes \(\in\) \{(1, 1, 1), (2, 2, 2)\}, and padding sizes \(\in\) \{(0, 0, 0), (1, 1, 1), (2, 2, 2), (0, 1, 2)\}, respectively.

\item \textbf{M2N2:} We vary the values of the hyperparameters test-time learning rates \(\in\) \{1e-01, 1e-02, 1e-03, 1e-04\}, latent dims \(\in\) \{8, 16, 32, 64\}, and gammas \(\in\) \{0.1, 0.4, 0.7, 0.9\}, respectively..

\item \textbf{MTGFLOW:} We vary the values of the hyperparameters learning rates \(\in\) \{1e-1, 3e-2, 3e-3, 1e-4\}, flow blocks \(\in\) \{1, 2\}, and hidden sizes \(\in\) \{8, 32, 64\}, respectively..

\item \textbf{GDFlow (Ours):} We vary the values of the hyperparameters learning rates \(\in\) \{1e-1, 3e-2, 3e-3, 1e-4\}, q-quantiles \(\in\) \{0, 0.01, 0.03, 0.05, 0.07, 0.1\}, flow blocks \(\in\) \{1, 2\}, and hidden sizes \(\in\) \{8, 32, 64\}, respectively.
\end{itemize}

\subsection{Experimental Setup for Generalization Ability}
\label{sec:a2}
We summarize the hyperparameters used in our experiments to evaluate the generalization ability as follows:

\begin{itemize}
\item \textbf{DBSCAN-DTW:} We set min samples to 3 and eps to 40. Only the test set is used since no training is needed.

\item \textbf{SAE-DBSCAN:} We set the learning rate to 3e-3, min samples to 2, eps to 1.0, and sparsity to 0.7. The model consists of an encoder and a decoder, each with a single layer. The embedding size is the same as the window size.

\item \textbf{GAE-DBSCAN:} We set the learning rate to 3e-3, min samples to 2, and eps to 0.8. The model consists of an encoder and a decoder, each with a single GRU layer. The embedding size is the same as the window size.

\item \textbf{DCEC:} We set the learning rate to 1e-3. The K-means parameter \( k \) is set to 2. The kernel size is \{2, 2, 1\}, the stride is \{1, 1, 1\}, and the padding is \{0,0,0\}. The encoder consists of layers \{32, 64, 128\}, and the decoder consists of layers \{128, 64, 32\}.

\item \textbf{M2N2:} We set the latent dimension to 16, the test-time learning rate to 0.1, and gamma to 0.4. We use a learning rate of 1e-3 and train for 10 epochs. The stride is the same as the window size, which is different from other models. No weight decay is used.

\item \textbf{MTGFLOW:} The model consists of 1 flow {block} with a batch size of 256. We set the stride to 1, and the train split to 0.8. The learning rate is 3e-3, and the number of epochs is 10. The hidden size is set to 32.

\item \textbf{GDFlow (Ours):} The model consists of 1 flow {block} with a batch size of 256. We set the stride to 1, and the train split to 0.8. The learning rate is 3e-3, and the number of epochs is 10. The hidden size is set to 32. The q-quantile values are set to \{0.01, 0.05, 0.07, 0.1\} for the {four} deceleration datasets.
\end{itemize}

\subsection{Experimental Setup for Benchmark Datasets}
\label{sec:a3}
The performance of \textbf{THOC}, \textbf{AT}, and \textbf{M2N2} follows the experimental results from ~\cite{kim2024model}, and thus the experimental settings are assumed to be the same as those in the referenced paper. The hyperparameter settings for other NF-based models and \fraName, covering the four datasets SMD (Machine-1-4 and Machine-2-1), MSL (P-15), and SMAP (T-3), are as follows:

\begin{itemize}

\item \textbf{GANF and MTGFLOW:} The model consists of 1 flow {block} with a batch size of 512. We set the window size to 60, the stride to 10, and the train split to 0.8. The learning rate is 2e-3, and the number of epochs is 40. The hidden size is set to 32. These settings are used for experiments on the {four} benchmark datasets.

\item \textbf{GDFlow (Ours):} The model consists of 1 flow {block} with a batch size of 256. We set the window size to \(\in\) \{60, 20, 10, 60\}, the stride to \(\in\) \{10, 1, 10, 10\}, and the train split to 1.0, meaning no validation set was used. The learning rate is \(\in\) \{3e-2, 3e-3, 1e-1, 1e-4\}, and the number of epochs is 20. The hidden size is \(\in\) \{8, 64, 32, 8\}. The q-quantile values are \(\in\) \{0.05, 0.01, 0.07, 0.1\}. These settings are used for experiments on the {four} benchmark datasets.

\end{itemize}

\section{Benchmark datasets}
\label{sec:bench}
The SMD dataset includes five weeks of data from 28 distinct server machines with 38-dimensional sensor inputs. For our experiments, we selected two specific server machines (Machine-1-4 and Machine-2-1) due to their distribution shift problems. The SMAP and MSL datasets, derived from spacecraft monitoring systems, also played a key role in our evaluation. The SMAP dataset comprises monitoring data from 28 unique machines with 55 telemetry channels, while the MSL dataset includes data from 19 unique machines with 27 telemetry channels. We selected data from two specific machines with distribution shifts, MSL (P-15) and SMAP (T-3), for our experiments.

\section{Comparison of Memory Efficiency and Computational Cost between NCDE and RNN}
\label{sec:mem}
For RNN models, the required memory is \(O(T \times h)\), where \(T\) is the sequence length, and \(h\) is the dimension of the hidden state. In contrast, NCDE requires only \(O(L + h)\) memory, where \(L\) is the integral time space \(T_1 - T_0\), and \(h\) is the size of the vector field. Since we use two NCDEs, our model requires \(O(2L + h_{f1} + h_{f2})\) memory. Unlike the memory requirements of RNNs, NCDE requires significantly less memory, making it not only performance-efficient but also cost-efficient.

Additionally, the ODE solvers used in NCDE address the slow convergence of traditional SDE {solvers,} and unlike RNNs that converge monotonically, they model dynamics over \([T_0, T_1]\), enabling faster convergence and fewer discretizations of the solver~\cite{finlay2020train, rubanova2019latent, jhin2022exit}. Thus, \fraName, as an NCDE-based model, can perform matrix-vector multiplication without large computational costs~\cite{kidger2020neural, jhin2024attentive}. This efficiency helps mitigate the increasing computational load as driving data accumulates in electric vehicles, making it suitable for real-world industrial deployment.

\end{document}